\documentclass[twoside,leqno,twocolumn]{article}

\usepackage[letterpaper]{geometry}

\usepackage{ltexpprt}
\usepackage{hyperref}

\usepackage{inconsolata}
\usepackage{multirow}
\usepackage{amsmath} % zxc
\usepackage{amsfonts} % zxc
\usepackage{graphicx} % zxc
\usepackage{float} % zxc
\usepackage{subfigure} % zxc
\usepackage{overpic} % zxc
\usepackage{bbm} % zxc

\begin{document}

\newcommand\relatedversion{}
\renewcommand\relatedversion{\thanks{The full version of the paper can be accessed at \protect\url{https://arxiv.org/abs/1902.09310}}} % Replace URL with link to full paper or comment out this line

%\setcounter{chapter}{2} % If you are doing your chapter as chapter one,
%\setcounter{section}{3} % comment these two lines out.

% \title{Feature Enhanced Zero-Shot Stance Detection via Contrastive Learning}

% \author{
%     Xuechen Zhao\thanks{National University of Defense Technology.}
%     \and Jiaying Zou\thanks{National University of Defense Technology.}
% }

\title{Zero-shot stance detection based on cross-domain \\feature enhancement by contrastive learning}
    \author{Xuechen Zhao\thanks{School of Computer, National University of Defense Technology, Changsha, China. \{zhaoxuechen, zoujiaying20, zhangzhong, xiefeng, binzhou, leitian129\}@nudt.edu.cn}
    \and Jiaying Zou\footnotemark[1]
    \and Zhong Zhang\footnotemark[1]
    \and Feng Xie\footnotemark[1]
    \and Bin Zhou\footnotemark[1]\hspace{2mm}\thanks{Corresponding Author, Key Lab. of Software Engineering for Complex Systems, Changsha, China.}
    \and Lei Tian\footnotemark[1]
}

\date{}

\maketitle

% Copyright Statement
% When submitting your final paper to a SIAM proceedings, it is requested that you include
% the appropriate copyright in the footer of the paper.  The copyright added should be
% consistent with the copyright selected on the copyright form submitted with the paper.
% Please note that "20XX" should be changed to the year of the meeting.

% Default Copyright Statement
% \fancyfoot[R]{\scriptsize{Copyright \textcopyright\ 20XX by SIAM\\
% Unauthorized reproduction of this article is prohibited}}

% Depending on which copyright you agree to when you sign the copyright form, the copyright
% can be changed to one of the following after commenting out the default copyright statement
% above.

%\fancyfoot[R]{\scriptsize{Copyright \textcopyright\ 20XX\\
%Copyright for this paper is retained by authors}}

%\fancyfoot[R]{\scriptsize{Copyright \textcopyright\ 20XX\\
%Copyright retained by principal author's organization}}

%\pagenumbering{arabic}
%\setcounter{page}{1}%Leave this line commented out.

\begin{abstract} \small\baselineskip=9pt 
Zero-shot stance detection is challenging because it requires detecting the stance of previously unseen targets in the inference phase. The ability to learn transferable target-invariant features is critical for zero-shot stance detection. In this work, we propose a stance detection approach that can efficiently adapt to unseen targets, the core of which is to capture target-invariant syntactic expression patterns as transferable knowledge. Specifically, we first augment the data by masking the topic words of sentences, and then feed the augmented data to an unsupervised contrastive learning module to capture transferable features. Then, to fit a specific target, we encode the raw texts as target-specific features. Finally, we adopt an attention mechanism, which combines syntactic expression patterns with target-specific features to obtain enhanced features for predicting previously unseen targets. Experiments demonstrate that our model outperforms competitive baselines on four benchmark datasets. 
\end{abstract}

\section{Introduction}

The goal of stance detection is to automatically identify the attitude or stance (e.g., Favor, Against, or Neutral) expressed in a text towards a specific target or topic\footnote{In the paper, we will use the terms: target and topic interchangeably.} \cite{augenstein2016stance, mohammad2016semeval, jang2018explaining}. The traditional target-specific stance detection assumes that the training and testing data belonged to the same target \cite{kuccuk2020stance}. However, due to the continuous emergence of unseen targets, collecting data on all targets for training is infeasible in practice. Moreover, it is expensive to obtain high-quality labels for a new target \cite{mohammad2016semeval}. Therefore, the study of zero-shot stance detection for unseen targets goes beyond the target-specific task and can help to predict stance more flexibly.

For the zero-shot stance detection task, some existing approaches try to improve the model's predictive ability for unseen targets by employing attention mechanisms \cite{allaway2020zero, xu2018cross} or fusing external knowledge \cite{liu2021enhancing}. However, transferring knowledge directly from a specific target to an unseen target is often limited in its predictive effectiveness due to the coupling of target-specific features. \cite{allaway2021adversarial, wei2019modeling} use adversarial learning to guide the model to learn target-invariant features via discriminators, which may lead to degraded prediction performance in the unbalanced distribution of targets. \cite{liang2022zero} capture the target-invariant features by identifying the stance feature categories and supervised contrastive learning, so their model achieves a better generalization capability.  However, the data needs to be tagged with soft labels by pretext tasks, which increases the complexity of the model and brings some noise to the data. Target-specific features are directly related to a specific target, while target-invariant features are generic and transferable, regardless of their targets. Consequently, it is crucial to distinguish these two features when predicting the stance of a text on unseen targets.

\begin{table*}
\centering
\setlength{\abovecaptionskip}{0.2cm} %调整标题上方的距离 
\renewcommand\arraystretch{1}
\setlength{\tabcolsep}{1pt}
\begin{tabular}{|clcc|}
\hline
\textbf{Example 1:}&{\qquad\qquad\textbf{Target}: Climate Change is a Real Concern} & \multicolumn{2}{r|}{\textbf{Stance}: Favor\qquad\qquad\qquad\qquad}\\
\multicolumn{4}{|p{16cm}|}{\textbf{Sentence}: Today Europe is breaking heat records, while Asia is breaking the lowest temperature records! Should we not be concerned?} \\
\multicolumn{4}{|p{16cm}|}{\textbf{Masked}: Today Europe is [MASK], while Asia is [MASK]! Should we not be [MASK]?} \\
\hline
\textbf{Example 2:}&{\qquad\qquad\textbf{Target}: Feminist Movement } & \multicolumn{2}{r|}{\textbf{Stance}: Favor\qquad\qquad\qquad\qquad}\\
\multicolumn{4}{|p{16cm}|}{\textbf{Sentence}: When they say men look at women like a piece of meat, what do they even mean, they want to cook \& eat her?} \\
\multicolumn{4}{|p{16cm}|}{\textbf{Masked}: When they say men look at [MASK] like [MASK] what do they even mean, they want to [MASK]?} \\
\hline
\end{tabular}
\caption{Examples of syntactic expression patterns.}
\label{tab:sample}
\end{table*} 

Both linguistic and psychological fields divide language into two aspects of representation: 1) syntactic representation and 2) semantic representation. The former reflects the form of languages, such as word morphology and sentence structure, and is the external representation of language; the latter is the concept and proposition, denoting the meaning referred to by the form of language, which is abstract and is the internal representation of language \cite{huiyang2018}. Meanwhile, Event-related Potentials (ERP) interaction theory \cite{carreiras2004line} suggests that text semantics is a fusion of syntactic and semantic representations, where syntax and semantics interact to complete the process of sentence comprehension and expression jointly. As shown in  Table \ref{tab:sample}, it is possible to use the same or similar syntactic expression patterns even for sentences with different targets, i.e., although the targets of Example 1 and Example 2 are distinct, they both use rhetorical question expression patterns, so these syntactic expression patterns are target-invariant. The target-invariant syntactic expression patterns and the target-specific features jointly determine the sentence's meaning. Inspired by this, we acquire syntactic expression patterns, which are naturally target-invariant and have an important impact on semantics. Furthermore, these syntactic expression patterns can combine with target-specific features to effectively predict the text stances.

More concretely, we propose a \underline{F}eature \underline{E}nhanced Zero-shot Stance Detection Model via \underline{C}ontrastive \underline{L}earning (\textbf{FECL}). First, we capture syntactic expression patterns via contrastive learning as transferable features. Second, based on supervised learning, we fully use the labeled data to learn semantic information as target-specific features. At last, a feature fusion module fuses the target-invariant and the target-specific features to achieve the cross-target prediction capability.

The main contributions of this paper can be summarized as follows:
\begin{itemize}
    \item We explore a novel self-supervised feature learning scheme. The scheme augments the raw texts by masking their topic words and then adopts contrastive learning to capture syntactic expression patterns as a bridge for knowledge transferring.
    \item We model the text stance expression into two parts: syntactic and semantic expression, and adopt an attention-based feature fusion mechanism that considers both target-invariant and target-specific features of texts. This mechanism improves the quality of feature representation and allows the model to handle prediction tasks such as zero-shot, cross-target, etc.
    \item Extensive experiments on four benchmark datasets show that the proposed model performs well on the zero-shot stance detection task. We also extend the model to few-shot and cross-target stance detection tasks, demonstrating the superiority and generalization of our approach.
\end{itemize}

\section{Related Work}
\paragraph{Stance Detection.}
Stance detection aims to examine the attitude of a text towards a given target \cite{hardalov2021survey}. It is widely applied in argument mining \cite{sirrianni2020agreement}, fake news/rumor detection \cite{kotonya2019gradual}, fact-checking  \cite{thorne2018fever}, and epidemic trend prediction  \cite{glandt2021stance}. Recently, with the rapid development of social media, various unseen targets have emerged, which also brings new challenges to stance detection. Therefore, cross-target and zero-shot stance detection have received extensive attention. Cross-target stance detection is training on source targets and adapting to unseen but correlated targets \cite{liang2021target}. While zero-shot stance detection aims to train on multiple targets with labels and then automatically identify an unseen target \cite{liang2022zero}. Their core ideas are learning target-invariant features and transferring them to unseen targets. \cite{augenstein2016stance} modeled the features of unseen targets via BiCondition LSTM. \cite{xu2018cross} extracted shared features based on a self-attention neural network. \cite{allaway2021adversarial} applied a target-specific stance dataset \cite{mohammad2016semeval} to zero-shot stance detection and used adversarial learning to capture target-invariant features. \cite{liu2021enhancing} proposed a Bert-based \cite{devlin2018bert} commonsense knowledge-enhanced graph model to zero-shot stance detection. These works try to extract transferable features from source targets and apply them to unseen targets but ignore the most basic syntactic expression patterns, which is a natural target-invariant and an essential factor that affects semantics and can effectively detect stance with the features of specific targets.

\paragraph{Contrastive Learning.}
Contrastive learning is a self-supervised representation learning method initially proposed in computer vision. It aims to learn distinct representations by pulling semantically close neighbors together and pushing non-neighbors away  \cite{hadsell2006dimensionality}. Recent studies have attempted to aid or enhance the performance of natural language processing tasks by incorporating contrastive learning  \cite{mohtarami2019contrastive,gao2021simcse,cai2020group}. Contrastive learning has proven comparable to supervised learning in many domains and effectively improves the quality of feature learning. However, most traditional stance detection algorithms only use supervised stance label information and inadequately utilize the richer information in the unlabeled text. Some researches \cite{liang2022zero, liang2022jointcl} began to apply contrastive learning to stance detection to improve the feature representation ability of the model. Therefore, our proposed model uses both the existing data labels and syntactic expression patterns to learn transferable features from source targets, thus improving the performance of stance detection for unseen targets.

\section{Methodology}

\begin{figure*}
    \setlength{\abovecaptionskip}{0cm} %调整标题上方的距离 
    \centering
    \includegraphics[width=0.9\linewidth]{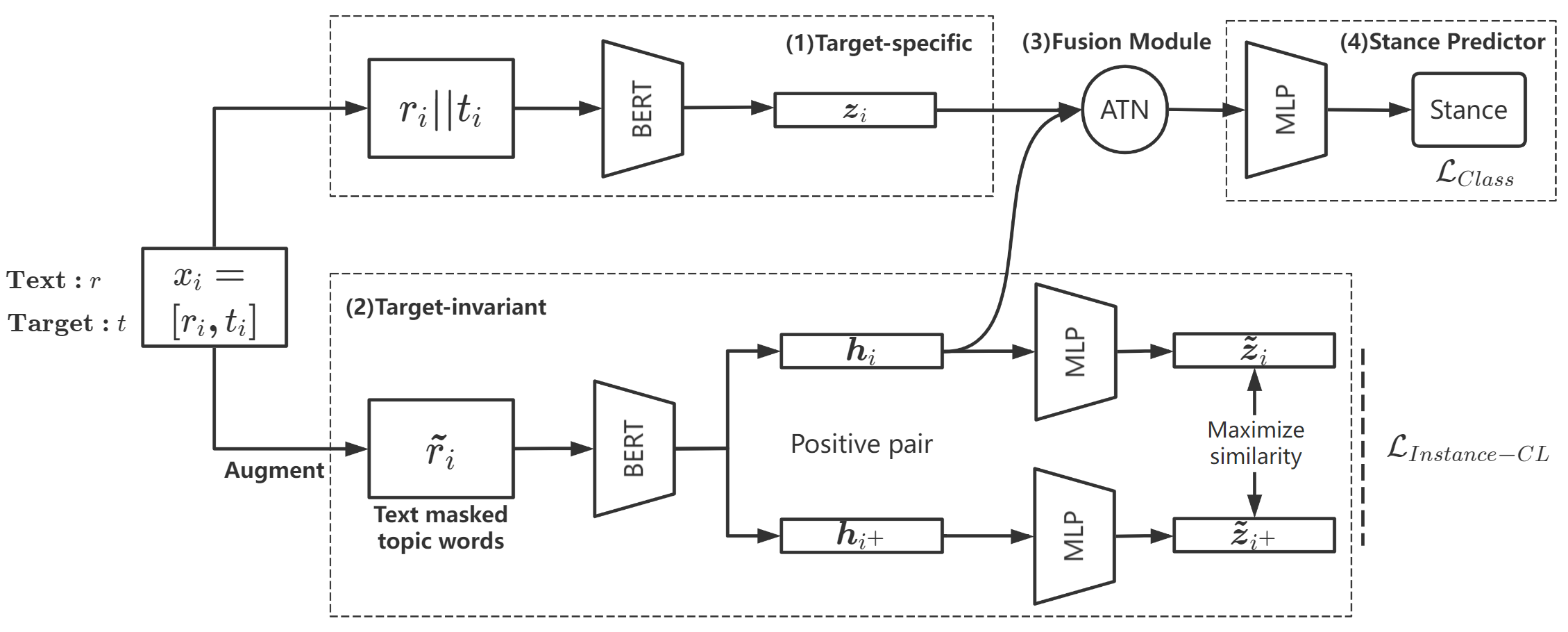}
    \caption{The architecture of the proposed EFCL model.}
    \label{fig:framework}
\end{figure*}

\subsection{Task Description}
Given a set of annotated instances toward some targets $\mathcal{D}_s = \{x_i^s=(r_i^s, t_i^s, y_i^s)\}_{i=1}^{N_s}$ and a set of unlabeled instances toward an unseen target $\mathcal{D}_d=\{x_i^d=(r_i^d, t^d)\}_{i=1}^{N_d}$, where $r_i^s$ and $r_i^d$ are the texts of instances in the training and testing sets, respectively, $t_i^s$(there may be many different targets in the training set) and $t^d$ are the target of instances in the training and testing sets, respectively, $y_i^s$ is the stance label of an annotated instance towards source target, $N_s$ and $N_d$ are the number of instances toward the source target and unseen target, respectively. The zero-shot stance detection aims to train the model from the training set and then generalize the model to the testing set where the target has never been seen to predict the stance.

\subsection{Data Augmentation}
Learning target-invariant features is the key to enhance transferable ability. This feature is mainly related to syntactic expression patterns, so we weaken the semantic content of each sentence and force our model to pay more attention to the part of syntactic expression patterns by masking the topic words. In this work, we deploy the latent Dirichlet allocation (LDA) \cite{blei2003latent} to obtain the topic words for each target. For each target, we generate T topics. Each topic contains K keywords, i.e., each target has T$\times$K keywords. Finally, we filter the duplicate keywords and the remaining ones as the candidates to be masked from sentences. Then, we formally represented the training and testing sets as $\mathcal{D}_s=\{x_i^s=(r_i^s,\tilde{r}_i^s,t_i^s,y_i^s)\}_{i=1}^{N_s}$ and $\mathcal{D}_d=\{x_i^d=(r_i^d,\tilde{r}_i^d,t^d)\}_{i=1}^{N_d}$, respectively, where $\tilde{r}_i$ is the augmented sentence that was masked topic words.

\subsection{The Proposed Framework}
In this section, we present our proposed FECL framework for zero-shot stance detection. As demonstrated in Figure \ref{fig:framework}, the proposed FECL framework includes four modules: 1) text encoder for specific targets (target-specific features learning). 2) syntactic expression patterns leaner based on contrastive learning (target-invariant features learning). 3) fusion module of target-specific features and target-invariant features based on attention mechanism. 4) text stance predictor. Regarding target adaptation, contrastive learning captures the underlying syntactic  expression patterns, which are target-invariant and can be transferred to an unseen target.

\subsubsection{Target-specific Features Representation}
This module focuses on the high-level semantics of sentences toward a target. Here, the sentence  $r=\{w_i\}_{i=1}^n$ towards a target $t$ is composed of $n$ words. The pre-trained BERT \cite{devlin2018bert} can be fine-tuned then as an encoder for diverse downstream tasks. To take full advantage of contextual information, we jointly embed target and text using BERT. 
\begin{equation}
    \boldsymbol{z},\mathbf{Z}=\textrm{BERT}([CLS]t[SEP]r[SEP])
\end{equation}
where, [CLS] is a special token that be used as the aggregate representation for the overall semantic context, while [SEP] is the special separator token. We obtain the $d_m$-dimensional hidden representation $\boldsymbol{z}\in\mathbb{R}^{d_m}$ of the token "[CLS]" and the feature matrix $\mathbf{Z}\in\mathbb{R}^{n\times{d_m}}$ of all the words forming the sentence $r$ at the last hidden layer.

\subsubsection{Syntactic Expression Patterns Learning via Contrastive Learning}
To learn target-invariant transferable features, we mainly focus on the syntactic expression patterns, formulate them as self-supervised representation learning problems, and use a contrastive learning algorithm to solve them. The success of self-supervised contrastive learning inspires the algorithm in visual representations \cite{chen2020simple,he2020momentum}. Here we only use the masked sentence 
$\tilde{r}$ for training. We also use a BERT model as the encoder $f_\theta(\cdot)$ to extract the representations of instances. Each instance is constructed as "[CLS]$\tilde{r}$[SEP]" and then fed to the encoder to obtain the hidden layer representation $h=\mathbb{R}^{d_m}$ of the token "[CLS]".

The key to apply contrastive learning on NLP is how to construct positive pairs of instances \cite{pan2021improved}. We follow \cite{gao2021simcse}, which notes that "even deleting one word would hurt performance and none of the discrete augmentations outperforms dropout noise". Hence, we adopt dropout acts as minimal "data augmentation" of hidden representations, which can maximize the retention of the original syntactic and semantic information. We denote $\boldsymbol{h}_{i}=f_\theta(\tilde{r},m)$ where $m$ is a random mask for dropout. Given an input mini-batch of data $\mathcal{B}=\{\boldsymbol{x_i}\}_{i=1}^{N_b}$,  $N_b$ is the size of the mini-batch, we simply feed the same instance to the encoder twice and get two embeddings $\boldsymbol{h}_i,\boldsymbol{h}_{i'}$ with different dropout masks $m$ and $m'$. Then, the mini-batch for contrastive learning is doubled to $\mathcal{B}'$, which size is $2N_b$. We refer to $\{\boldsymbol{h}_i,\boldsymbol{h}_{i'}\}$ as a postive pair, while treat the other $2N_{b}-2$ samples in $\mathcal{B}'$ as negative instances regarding this positive pair. Additionally, we define a neural network projection head $\boldsymbol{\tilde{z}}_i=g_\phi(\boldsymbol{h_i}) = W^{(2)}\sigma(W^{(1)}h_i)$ that maps the representations to the space for computing the contrastive loss, where $\sigma(\cdot)$ is a nonlinear activation function (i.e., ReLU). For each 
instance, the contrastive loss function is formulated as:
\begin{equation}
    \ell_i = -log\frac{e^{sim(\tilde{z}_i,\tilde{z}_{i'})/\tau}}{\sum_{j=1}^{2N_b}\mathbbm{1}_{j\neq{i}}\cdot{e}^{sim(\tilde{z}_i,\tilde{z}_{j})/\tau}}
\end{equation}
where $\mathbbm{1}_{j\neq{i}}$ is an indicator function, $sim(\boldsymbol{u},\boldsymbol{v})=\boldsymbol{u}^T\boldsymbol{v}/\Vert\boldsymbol{u}\Vert\Vert\boldsymbol{v}\Vert$ represents the cosine similarity of normalized outputs u and v, $\tau$ is a temperature parameter that controls the penalty intensity for hard sample in contrastive learning \cite{wang2021understanding}. The contrastive learning loss for each mini-batch is formulated as:
\begin{equation}
    \mathcal{L}_{Instance-CL}=\frac{-1}{2N_b}\sum_{i\in\mathcal{B}'}{\ell_i}
\end{equation}

\subsubsection{Feature Fusion based on Attention Mechanism}
The stance features are represented by the joint of target-invariant and target-specific. But, directly concatenating the two parts may lose some knowledge in the encoding \cite{zhang2020enhancing}. Inspired by  \cite{zhang2019aspect}, we adopt a retrieval-based attention mechanism to fuse the features of syntactic expression patterns and target-specific features, where we select any representation $\boldsymbol{h}_i$ in the positive pair of the contrastive learning as the target-invariant features:
\begin{equation}
    \alpha_{ij}=\frac{\textrm{exp}(\beta_{ij})}{\sum_{t=1}^{n}\textrm{exp}(\beta_{it})}
\end{equation}
\begin{equation}
    \beta_{ij}=(\mathbf{W}_{q}\boldsymbol{h}_i)^{T}(\mathbf{W}_{k}\mathbf{Z}_{ij})
\end{equation} 
where $a_{ij}$ is the attention of the $i$-th instance semantic expression pattern feature on the $j$-th word, and $\mathbf{Z}_{ij}$ is the hidden representation of the $j$-th word from the $i$-th sample, $T$ represents matrix transposition, $\textrm{W}_q,\textrm{W}_k\in\mathbb{R}^{d_s\times{d_m}}$ are parameters to be learned, $d_s$ is the dimensionality of the feature after mapping. Furthermore, the final stance feature $\boldsymbol{f}_i\in\mathbb{R}^{2d_m+d_s}$ for each input instance can be formulated as:
\begin{equation}
    \boldsymbol{f}_i=\boldsymbol{h}_i\Vert\boldsymbol{z}_i\Vert\sum_{t=1}^{n}\alpha_{it}(\mathbf{W}_{v}\mathbf{Z}_{it})
\end{equation} 
where || is the concatenation operation, and $\mathbf{W}_v\in\mathbb{R}^{d_s\times{d_m}}$ denotes trainable parameters.

Furthermore, we adopt a fully-connected layer with softmax normalization to yield a probability distribution of stance prediction: 
\begin{equation}
    \hat{y}=\textrm{softmax}(\mathbf{W}_{o}\boldsymbol{f}_i+\boldsymbol{b}_o)
\end{equation}
where $\hat{y}\in\mathbb{R}^{d_p}$ is the predicted probability of the input instance $x_i$, $d_p$ is the dimensionality of the stance labels, $\mathbf{W}_o\in\mathbb{R}^{d_p\times{(d_m+d_s)}}$ and $\boldsymbol{b}_o\in\mathbb{R}^{b_p}$ are trainable parameters.

Finally, we train the classifier by the cross-entropy loss between the predicted distribution $\hat{\boldsymbol{y}}$ and ground-truth distribution $\boldsymbol{y}$ of every instance:
\begin{equation}
    \mathcal{L}_{cls}=-\sum_{i=1}^{N_b}\sum_{j=1}^{d_p}y_i^jlog\hat{y}_i^j
\end{equation}

\subsubsection{Learning Objective}
The learning objective of our proposed model is to train the model by jointly optimizing a supervised loss of stance classification $\mathcal{L}_{cls}$ with a contrastive loss $\mathcal{L}_{Instance-CL}$. The overall loss $\mathcal{L}$ can be formulated as the sum of three losses:
\begin{equation}
    \mathcal{L}=\mathcal{L}_{cls}+\eta\mathcal{L}_{Instance-CL}+\lambda\Vert\Theta\Vert^2
\end{equation}
where $\eta$ is a tuned hyper-parameter, $\Theta$ represents all trainable parameters in the model, and $\lambda$ denotes the $L_2$-regulateion coefficient.

\begin{table}[htbp]
\renewcommand\arraystretch{1}
\setlength\tabcolsep{4pt}
% \vspace{-0.8cm} 
\centering
\begin{tabular}{cc|ccc}
\hline
\textbf{DataSet} & \textbf{Target}&\textbf{Favor}&\textbf{Against}&\textbf{Neutral}\\
\hline
\multirow{7}*{\textbf{SEM16}} & DT & 148 & 299 &260 \\
& HC & 163 &565&256 \\
& FM & 268 &511&170 \\ 
& LA & 167 &544&222 \\
& AT & 124 &464&145 \\
& CC & 335 &26&203 \\
& TP & 333 &452&460 \\ 
\hline
\multirow{4}*{\textbf{WT-WT}} 
& CA & 2469 & 518 & 5520\\
& CE & 773 & 253 & 947\\
& AC & 970 & 1969 & 3098\\
& AH & 1038 & 1106 & 2804\\
\hline
\multirow{4}*{\textbf{COVID-19}} & WA & 515 & 220 & 172\\
& SC & 430 & 102 & 85\\
& AF & 384 & 266 & 307\\
& SH & 151 & 201 & 396\\
\hline
\end{tabular}
\caption{Statistics of of SEM16, WT-WT and COVID-19 datasets.}
\label{tab:semcovid}
\end{table}

\begin{table}[htbp]
    \vspace{-0.2cm} 
    \centering
    \renewcommand\arraystretch{1}
    \begin{tabular}{c|ccc}
    \hline
    &\textbf{Train} & \textbf{Dev}&\textbf{Test}\\
    \hline
    \textbf{\#Examples} & 13477 & 2062 & 3006 \\
    \#Unique Comments & 1845 & 682 & 786 \\
    \#Zero-shot Topics & 4003 & 383 & 600 \\ 
    \#Few-shot Topics & 638 & 114 & 159 \\\hline
    \end{tabular}
    \caption{Statistics of VAST dataset.}
    \label{tab:vast}
\end{table}

\section{Experiments}
\subsection{Experimental Data}
\paragraph{SEM16 \cite{mohammad2016semeval}} SEM16 is a Twitter dataset for stance detection. The SEM16 contains 6 targets, namely Donald Trump (DT), Atheism (AT), Climate Change is a Real Concern (CC), Feminist Movement (FM), Hillary Clinton (HC), and  Legalization of Abortion (LA). Each tweet in the dataset contains a stance(Favor, Against, or Neutral) to a special target. To achieve zero-shot stance detection, we select all the samples from one of the targets as the testing set, and the samples of the other five targets are randomly divided into training sets and validation sets at a ratio of 85:15.

We further selected two pairs of targets (DT-HC and FM-LA) to validate the proposed model's cross-target stance detection capability. These targets constitute the US political domain and the female domain. Following \cite{wei2019modeling}, We split the annotated data of the destination target to obtain development and testing set with 3:7. The distribution of the dataset is shown in Table \ref{tab:semcovid}.

\paragraph{VAST \cite{allaway2020zero}} VAST is a public dataset for zero-shot and few-shot stance detection, which consists of a large range of targets. For zero-shot stance detection, instances in the training and validation sets will not appear in the testing set. However, there are a small number of instances in the validation and testing sets with the same target as the training set for a few-shot environment. The statistics of the dataset are shown in Table \ref{tab:vast}.

\paragraph{WT-WT \cite{wtwt}}
WT-WT is a financial dataset to detect the stance of M\&A operations between companies. We use four target pairs in the healthcare area from WT-WT, including CVS\_AET (CA), CI\_ESRX (CE), ANTM\_CI (AC), and AET\_HUM (AH). Each tweet from WT-WT is labeled with one of four targets (Refute, Comment, Support, and Unrelated), where Refute corresponds to Against, Comment corresponds to Neural and Support corresponds to Favor. To be consistent with other dataset classifications, we removed instances labeled as unrelated. We sequentially select all the samples from one of the targets as the testing set, and the cases of the other three targets are randomly divided into training sets and validation sets at a ratio of 85:15. The distribution of the dataset is shown in Table \ref{tab:semcovid}.

\paragraph{COVID-19 \cite{glandt2021stance}} The dataset contains 4 targets on COVID-19 policy, and since some of the original tweets were unavailable due to privacy concerns, we obtained a total of 3229 tweets. The targets of this dataset consist of Wearing a Face Mask (WA), Keeping Schools Closed (SC), Anthony S. Fauci, M.D. (AF), and Stay at Home Orders (SH). Each instance contains a stance label for a specific target: In-Favor, Against, or Neither. We adopt one target as the testing set, the other three as the training set, and randomly select 15\% from the training set as the develepment data. The statistics of the dataset are shown in Table \ref{tab:semcovid}.

\subsection{Baselines}
We evaluate and compare our model with several robust baselines. These include BiCond \cite{augenstein2016stance}, a neural network-based model. CrossNet \cite{xu2018cross}, an attention mechanism-based model. SEKT \cite{zhang2020enhancing}, a knowledge-based approach. TOAD \cite{allaway2021adversarial}, an adversarial learning-based model. TPDG \cite{liang2021target}, a graph neural network-based model. BERT \cite{devlin2018bert}, BERT-GCN \cite{liu2021enhancing}, BERT-based models. and PT-HCL \cite{liang2022zero}, a contrastive learning-based model.

Further, we designed several variants of our proposed model to analyze the impact brought by different components. "$\textrm{w/}\enspace{concat}$" denotes that the syntactic features and semantic features are fused by concatenating. "$\textrm{w/o} \enspace masktopic$" variant randomly masks 15\% of words in each sentence to generate masked sentences, and "$\textrm{w/o}\enspace\mathcal{L}_{Instance-CL}$" represents without using contrastive learning loss in the model training.

\subsection{Experimental Implementation}
All programs are implemented using Python 3.8.5 and PyTorch 1.9.1 with CUDA 11.1 in an Ubuntu server with an Nvidia 3090Ti GPU. 

\subsubsection{Evaluation Metrics}
For SEM16, following \cite{allaway2021adversarial}, we perform Macro-averaged F1 of Favor and Against in the zero-shot scenario. In cross-target stance detection, we perform the mean value of micro-averaged F1 and macro-averaged F1 of Favor and Against to alleviate the imbalance problem of target data. For VAST, following \cite{allaway2020zero}, we calculate Macro-averaged F1 values for each label to test the model's performance. For WT-WT and COVID-19, we also used Macro-averaged F1 of Support/Pro and Refute/Con to evaluate the performance of stance classification.

\begin{table*}[htbp]
\renewcommand\arraystretch{0.9}
\centering
\setlength\tabcolsep{2.8pt}
\begin{tabular}{c|c|ccccccc|cc}
\hline
\multicolumn{2}{c|}{\textbf{Model}} &BiCond&CrossNet&TOAD&BERT&BERT-GCN&TPDG&PT-HCL&\textbf{FECL}&$\textrm{w/}{concat}$\\
\hline
\multirow{6}*{SEM16}
&DT&30.5$^\ast$&35.6$^\Box$&49.5$^\ast$&40.1$^\ast$&42.3$^\Box$&47.3$^\Box$&\underline{50.1}&\textbf{51.6}&47.8\\
&HC&32.7$^\ast$&38.3$^\Box$&51.2$^\ast$&49.6$^\ast$&50.0$^\Box$&50.9$^\Box$&\underline{54.5}&\textbf{55.6}&53.1\\
&FM&40.6$^\ast$&41.7$^\Box$&54.1$^\ast$&41.9$^\ast$&44.3$^\Box$&53.6$^\Box$&\underline{54.6}&\textbf{55.4}&50.7\\
&LA&34.4$^\ast$&38.5$^\Box$&46.2$^\ast$&44.8$^\ast$&44.2$^\Box$&46.5$^\Box$&\underline{50.9}&\textbf{53.3}&49.1\\
&AT&31.0$^\ast$&39.7$^\Box$&46.1$^\ast$&55.2$^\ast$&53.6$^\Box$&48.7$^\Box$&\underline{56.5}&\textbf{57.3}&43.7\\
&CC&15.0$^\ast$&22.8$^\Box$&30.9$^\ast$&37.3$^\ast$&35.5$^\Box$&32.3$^\Box$&\underline{38.9}&\textbf{41.8}&38.6\\
\hline
\multirow{4}*{VAST}
&Pro&45.9$^\circ$&46.2$^\circ$&42.6$^\Box$&54.6$^\circ$&58.3$^\circ$&-&\textbf{61.7}&\underline{60.6}&59.4\\
&Con&47.5$^\circ$&43.4$^\circ$&36.7$^\Box$&58.4$^\circ$&60.6$^\circ$&-&\underline{63.5}&\textbf{67.3}&63.6\\
&Neu&34.9$^\circ$&40.4$^\circ$&43.8$^\Box$&85.3$^\circ$&86.9$^\circ$&-&\underline{89.6}&\textbf{89.8}&88.5\\
&All&42.7$^\circ$&43.4$^\circ$&41.0$^\Box$&66.1$^\circ$&68.6$^\circ$&-&\underline{71.6}&\textbf{72.5}&70.5\\
\hline
\multirow{4}*{WT-WT}
&AC&64.9$^\diamondsuit$&65.1$^\diamondsuit$&59.2&67.1$^\Box$&70.7$^\Box$&74.2$^\triangle$&\textbf{76.7}&\underline{76.1}&73.0\\
&AH&63.0$^\diamondsuit$&62.3$^\diamondsuit$&62.0&67.3$^\Box$&69.2$^\Box$&73.1$^\triangle$&\underline{76.3}&\textbf{76.4}&75.4\\
&CA&56.5$^\diamondsuit$&59.1$^\diamondsuit$&58.1&56.0$^\Box$&67.8$^\Box$&66.8$^\triangle$&\textbf{73.1}&\underline{72.6}&70.7\\
&CE&52.5$^\diamondsuit$&54.5$^\diamondsuit$&57.8&60.5$^\Box$&64.1$^\Box$&65.6$^\triangle$&\underline{69.2}&\textbf{70.5}&70.2\\
\hline
\multirow{4}*{COVID-19}&AF&26.7&41.3&40.1&\underline{47.5}&-&46.0&41.7&\textbf{50.7}&46.7\\
&SC&33.9&40.0&47.3&45.1&-&\underline{51.6}&44.7&\textbf{53.3}&49.1\\
&SH&19.3&40.4&42.0&39.7&-&37.3&\underline{53.3}&\textbf{54.9}&46.1\\
&WA&30.1&38.2&37.9&44.3&-&48.4&\textbf{58.8}&\underline{58.3}&54.8\\
\hline
\end{tabular}
\caption{Performance comparison of zero-shot stance detection on four datasets. Bold face indicates the best result of each column and underlined the second-best. The results with $\ast$ are retrieved from \cite{allaway2021adversarial}, with $\circ$ are retrieved from \cite{liu2021enhancing}, with $\Box$ are retrieved from \cite{liang2022zero}, with $\diamondsuit$ are retrieved from \cite{wtwt}, with $\triangle$ are retrieved from \cite{liang2021target}.}
\label{tab:zero}
\end{table*}

\subsubsection{Training Settings}
We use the pre-trained uncased BERT-base \cite{devlin2018bert} as the encoder with 768-dim embedding, the learning rate is 2e-5, and the coefficient of L2-regularization $\lambda$ is set to 1e-5. Adam is utilized as the optimizer. The mini-batch is set to 32. For contrastive learning loss, we set the hyper-parameters $ \tau=0.07 $, $ \eta=0.1 $ and dropout probablity $ p=0.1 $. We use LDA \cite{blei2003latent} to generate topic words for masking sentences, where T = 6 and K = 5. For BiCond and CrossNet, the word embeddings are initialized with the pre-trained 200-dimensional GloVe vectors  \cite{pennington2014glove}. The reported results are averaged scores of 10 runs to obtain statistically results. 

\subsection{Experimental Results}
\subsubsection{Main Experimental Results}
We report the main experimental results of zero-shot stance detection on four benchmark datasets in Table \ref{tab:zero}. We observe that our FECL outperforms the baseline models on most datasets, which verifies the effectiveness of our proposed approach in zero-shot stance detection task. 

Owing to the limitation of unseen destination target information, BiCond and CrossNet overall perform worst since they neither leverage target-specific contextual information nor learn transferable knowledge for the unseen target. Our model obtained excellent performance compared to the adversarial learning-based TOAD that performed poorly on VAST dataset, which indicates that it is difficult for adversarial learning to capture transferable information in the imbalanced distribution of targets. PT-HCL adopts the hierarchical supervised contrastive learning method, which has achieved good results, but the quality of pseudo-labels in its preprocessing may affect its prediction quality. At the same time, through contrastive learning, our FECL model utilizes the existing labels and further effectively mines the feature at the text structure level. It effectively learns the similarities and differences of sentences under different targets from semantic and syntactic aspects, improving performance. Similarly, BERT can employ rich semantic information, but it still performs poorly because it ignores the target-invariant stance expression on unseen targets. This shows that the self-supervised learning method can obtain the transferable features from source targets, and it effectively enhances the target-specific features for the previously unseen targets. Further, our FECL is significantly superior to the methods without contrastive learning in terms of all evaluation metrics. This demonstrates the significance and validity of our proposed contrastive learning approach in zero-shot stance detection.

\subsubsection{Task Generalization Ability Experiment}
We also conduct experiments in few-shot and cross-target scenarios to verify the effectiveness and generalization ability of our FECL.

First, we evaluate the stance detection performance of our model on VAST dataset in the few-shot scenario. The experimental results are shown in Table \ref{tab:few}. The FECL outperforms all the comparison models, verifying that our model has good performance and generalization ability in few-shot scenarios.

Second, we experiment with cross-target stance detection on SEM16 dataset, and the experimental results are shown in Table \ref{tab:cross}. Our FECL performs well on most dataset, demonstrating that our model can generalize to cross-target scenarios. However, the LA$\rightarrow$FM effect is slightly lacking, possibly due to only considering the syntactic expression pattern features of the source target in the training phase. These patterns can not cover the unseen target, the problem we will address in the future. BiCond and CrossNet performed the worst overall because they were coupled too much to the source target, and it was not easy to learn the transferable knowledge effectively. Although BERT can utilize rich semantic information, its performance is still poor since it dose not consider the distribution of features on unseen targets. The results show that simply modeling the transferable information from the source target to the unseen target is insufficient for stance prediction. In contrast, our FECL captures the target-invariant syntactic expression patterns, effectively improving cross-target stance detection performance. 

\subsubsection{Ablation Study}

To investigate the contribution of the different components of our proposed model, we conduct an ablation study on COVID-19 dataset and reported the results in Table \ref{tab:ablation}. We observe that the "$\textrm{w/o} \enspace topicmask$" variant performs poorly when masking some words based on randomness, suggesting that the choice of mask candidates is crucial for stance prediction. Effective mask candidates can remove the target information without destroying the syntactic expression patterns, which is beneficial to obtain target-invariant features. In addition, the "$\textrm{w/o} \enspace \mathcal{L}_{Instance-CL}$" variant without the contrastive learning loss also leads to performance drops, suggesting that the features obtained from contrastive learning can effectively enhance the learning quality. 

\subsubsection{Analysis of Contrastive Representation}
\begin{figure}[htbp]
    \centering
    \vspace{-0.2cm} 
    \setlength{\abovecaptionskip}{0cm} %调整标题上方的距离 
    \includegraphics[width=0.8\linewidth]{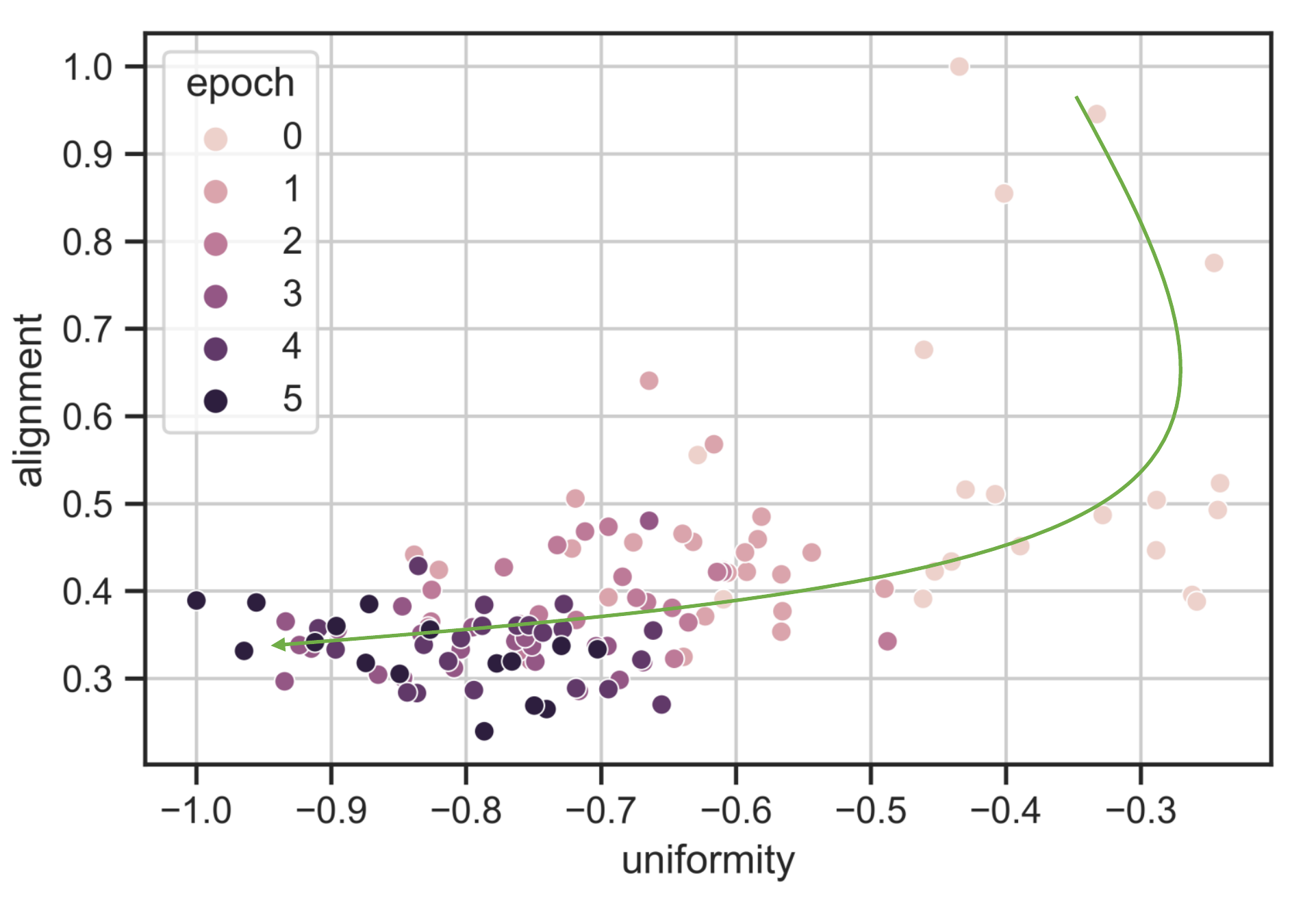}
    \caption{We visualize checkpoints every 5 training steps and the arrows indicate the training direction. The arrow indicates the training direction.}
    \label{fig:ali_uni}
\end{figure}

To further analyze the effectiveness of contrast learning in the proposed FECL, following \cite{wang2020understanding}, we use alignment and uniformity metrics to verify the model's performance. We take the checkpoints of our model every 5 steps during training and visualize the alignment and uniformity metrics in Figure \ref{fig:ali_uni}. As clearly shown, starting from pre-trained checkpoints, our model greatly improves uniformity and keeps a steady alignment. These results denote that the model learns better alignment and uniformity, which further facilitates the model to achieve better performance.

\begin{figure}[htbp]
    % \vspace{-0.2cm} 
    \setlength{\abovecaptionskip}{0cm} %调整标题上方的距离 
    \centering
    \subfigure[zero-shot.] {
        \label{fig:a}     
        \includegraphics[width=0.465\columnwidth]{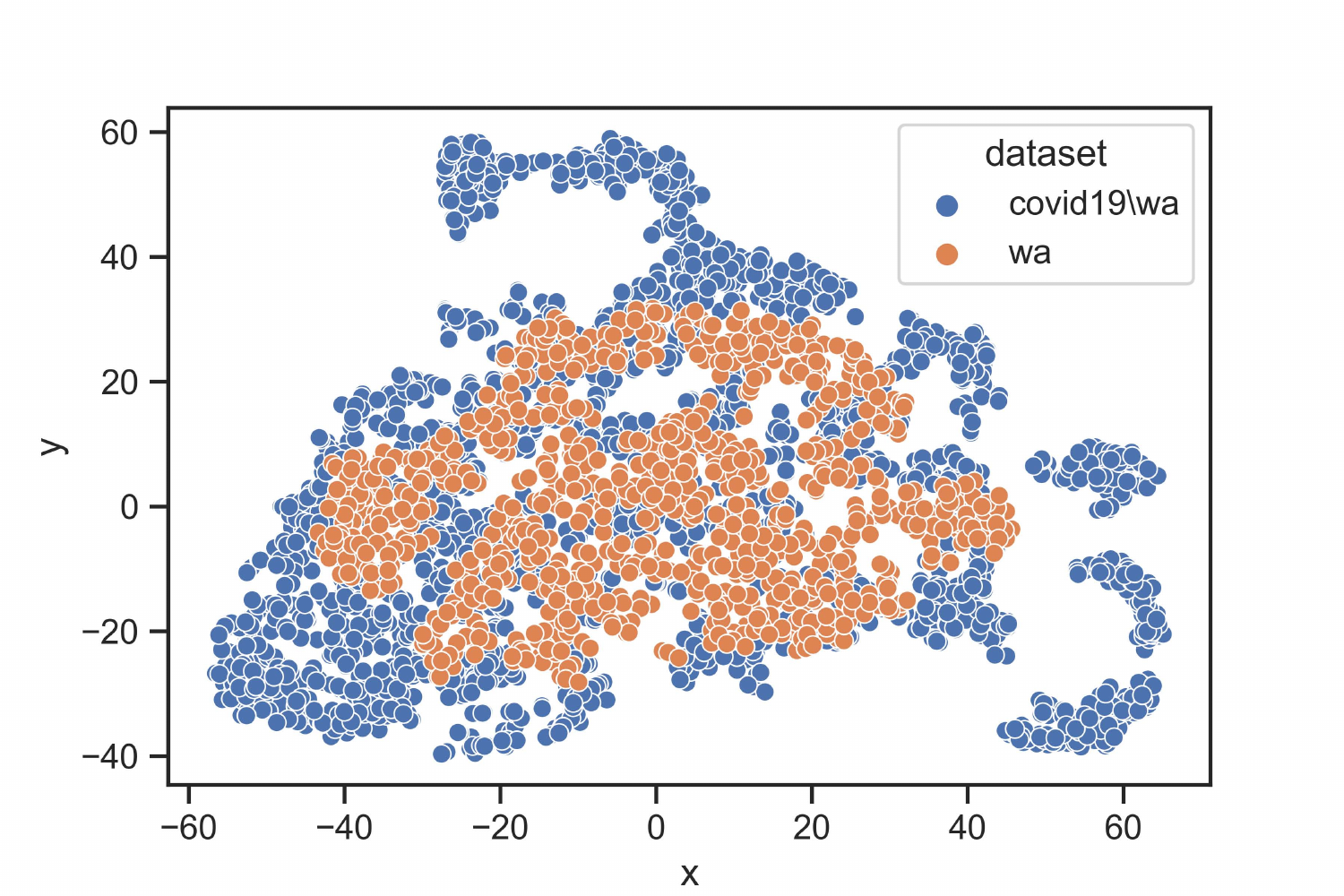}  
    }     
    \subfigure[cross-target.] { 
        \label{fig:b}     
        \includegraphics[width=0.465\columnwidth]{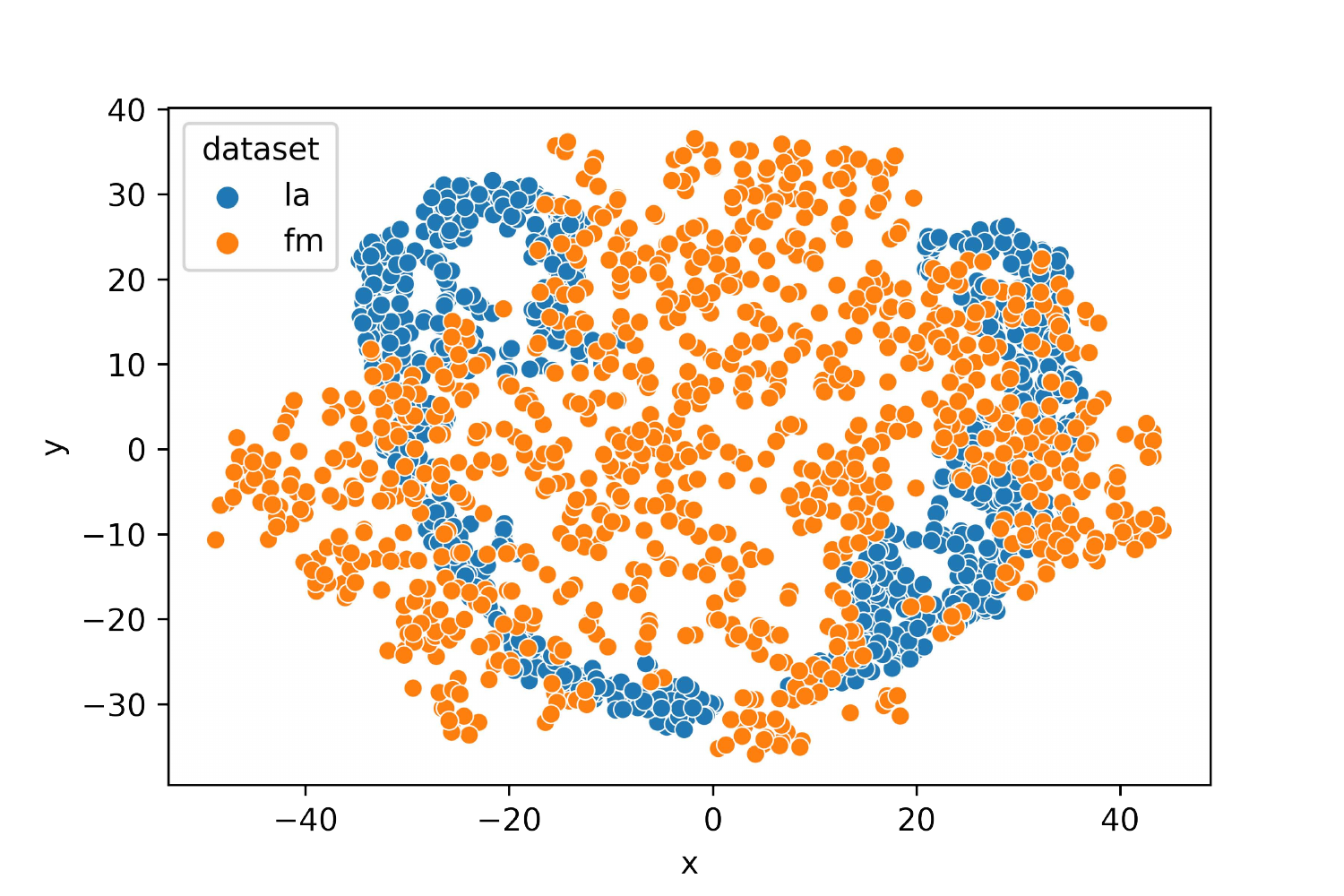}     
    }
    \caption{Visualization of syntactic expression pattern features learned by our FECL. The blue and orange colors indicate the distribution of the semantic expression pattern features of the training and testing datasets, respectively.}
    \label{fig.con_repr}
\end{figure}

To demonstrate the effectiveness of the learned contrastive representation, we use T-SNE \cite{van2008visualizing} visualization on zero-shot stance detection and cross-target stance detection, respectively. We show the results in Figure \ref{fig.con_repr}, the dots in the figure represent the syntactic expression pattern features learned by contrastive learning. Figure \ref{fig.con_repr}(a) shows that the target-invariant features retained in the training set (i.e., COVID-19 without WA) can effectively cover the testing set (i.e., WA), achieving better results in the zero-shot task. Figure \ref{fig.con_repr}(b) shows the contrastive representations of LA and FM in SEM16 dataset. Since the contrastive representation of FM can cover LA better, the cross-target task of FM$\rightarrow$LA fetches better results. In contrast, the expression pattern features in LA cannot cover FM well, so the results in the cross-target task of LA$\rightarrow$FM are weak. The results imply that the FECL can learn more reliable target-invariant features from the training due to the unsupervised contrast learning, which leads to better generalization to unseen targets and improves the performance of zero-shot stance detection.

\begin{table}[H]
    \vspace{-0.3cm}
    \renewcommand\arraystretch{1}
    \centering
    \begin{tabular}{c|cccc}
    \hline
    \multirow{2}*{\textbf{Model}} & \multicolumn{4}{c}{\textbf{VAST}}\\
    \cline{2-5}
    &\textbf{Pro}&\textbf{Con}&\textbf{Neu}&\textbf{All}\\
    \cline{2-5}
    \hline
    BiCond &45.4&46.3&25.9&39.2\\
    CrossNet &50.8&50.5&41.0&47.4\\
    SEKT &51.0&47.9&21.5&47.4\\
    BERT &54.4&59.7&79.6&64.6\\
    BERT-GCN &\underline{62.8}&63.4&83.0&69.7\\
    PT-HCL &62.3&\textbf{67.0}&\underline{84.3}&\underline{71.2}\\
    \hline
    FECL &\textbf{64.5}&\underline{65.3}&\textbf{87.0}&\textbf{72.3}\\
    \hline
    \end{tabular}
    \caption{Experimental results of few-shot scenario. Bold face indicates the best result of each column and underlined the second-best. The results of baselines are retrieved from  \cite{liang2022zero}.}
    \label{tab:few}
\end{table}

\begin{table}[H]
    % \vspace{-0.3cm} 
    \renewcommand\arraystretch{1}
    \centering
    \begin{tabular}{c|cccc}
    \hline
    \multirow{2}*{\textbf{Model}} & \multicolumn{4}{c}{\textbf{SEM16}}\\
    \cline{2-5}
    & {\textbf{\scriptsize{FM$\rightarrow$LA}}}&{\textbf{\scriptsize{LA$\rightarrow$FM}}}&{\textbf{\scriptsize{HC$\rightarrow$DT}}}&\textbf{\scriptsize{DT$\rightarrow$HC}}\\
    \cline{2-5}
    \hline
    BiCond &40.3&39.2&44.2&40.8\\
    CrossNet &44.2&43.1&46.1&41.8\\
    SEKT &53.6&51.3&47.7&42.0\\
    BERT &49.9&39.5&46.0&55.3\\
    TPDG &\underline{62.4}&\textbf{55.9}&51.0&\underline{57.6}\\
    PT-HCL &56.3&50.8&\underline{53.6}&56.5\\
    \hline
    FECL &\textbf{63.5}&\underline{52.1}&\textbf{54.6}&\textbf{59.4}\\
    \hline
    \end{tabular}
    \caption{Experimental results of cross-target scenario. Bold face indicates the best result of each column and underlined the second-best. "FM$\rightarrow$LA" means training on FM and testing on LA, etc. We calculated the results of PT-HCL according to the model proposed in \cite{liang2022zero}. The results of the other baselines are retrieved from \cite{liang2021target}.}
    \label{tab:cross}
\end{table}

\begin{table}[htbp]
    % \vspace{-0.2cm} 
    \renewcommand\arraystretch{1}
    \centering
    \begin{tabular}{c|cccc}
    \hline
    \multirow{2}*{\textbf{Model}} & \multicolumn{4}{c}{\textbf{COVID-19}}\\
    \cline{2-5}
    &\textbf{AF}&\textbf{SC}&\textbf{SH}&\textbf{WA}\\
    \cline{2-5}
    \hline
    $\textrm{w/o} \enspace topicmask$ &45.5&47.4&43.5&44.6\\
    $\textrm{w/o} \enspace \mathcal{L}_{Instance-CL}$ &46.4&45.2&42.9&46.3\\
    \hline
    FECL &\textbf{50.7}&\textbf{53.3}&\textbf{54.9}&\textbf{58.3}\\
    \hline
    \end{tabular}
    \caption{Experimental results of ablation study.}
    \label{tab:ablation}
\end{table}


\section{Conclusion}
In this paper, we propose a novel zero-shot stance detection model (FECL), which achieves stance prediction on unseen targets by fusing the target-invariant syntactic expression pattern features with the target-specific semantic features. Experimental results on four benchmark datasets show that our model exhibits state-of-the-art performance on many unseen targets. In addition, we extend the training to few-shot and cross-target scenarios with similarly good results, demonstrating the model's superior generalization ability.

\section*{Acknowledgements}
This work is supported by the Key R\&D Program of Guangdong  Province No.2019B010136003, the National Natural Science Foundation of China No. 62172428, 61732004, 61732022 and the Natural Science Foundation of Shandong Province of China No.ZR2020MF152.

\end{document}